%% file: main.tex
\documentclass[twoside,twocolumn,10pt]{article}

\usepackage{wscg}           
\RequirePackage{ifpdf}
\ifpdf
 \RequirePackage[pdftex]{graphicx}
 \RequirePackage[pdftex]{color}
\else
 \RequirePackage[dvips,draft]{graphicx}
 \RequirePackage[dvips]{color}
\fi
\usepackage{tikz}
\usetikzlibrary{shapes,arrows,shadows,decorations.markings}

\usepackage{float}
\usepackage{multirow}

\usepackage{glossaries}

\newif\ifanonymous
\anonymousfalse

\input{paper_gls.tex}

\title{A Framework for Visually Realistic Multi-robot Simulation in Natural Environment}

\ifanonymous
\usepackage[switch,pagewise]{lineno}
\linenumbers

\author{
\parbox{0.25\textwidth}{\centering
Anonymous Author\\[1mm]
author's affiliation\\
1st line of address\\
2nd line of address\\
Country (ZIP) code, City, State\\[1mm]
e@mail
}
\hspace{0.05\textwidth}
\parbox{0.25\textwidth}{\centering
Anonymous Author\\[1mm]
author's affiliation\\
1st line of address\\
2nd line of address\\
Country (ZIP) code, City, State\\[1mm]
e@mail
}
\hspace{0.05\textwidth}
\parbox{0.25\textwidth}{\centering
Anonymous Author\\[1mm]
author's affiliation\\
1st line of address\\
2nd line of address\\
Country (ZIP) code, City, State\\[1mm]
e@mail
}
}

\else

\usepackage{nopageno}

\author{
\parbox{0.4\textwidth}{\centering
Ori Ganoni\\[1mm]
Department of Computer Science\\
University of Canterbury\\
Christchurch, New Zealand\\[1mm]
ori.ganoni@pg.canterbury.ac.nz
}
\hspace{0.05\textwidth}
\parbox{0.4\textwidth}{\centering
Ramakrishnan Mukundan\\[1mm]
Department of Computer Science\\
University of Canterbury\\
Christchurch, New Zealand\\[1mm]
mukundan@canterbury.ac.nz
}
}

\fi

\usepackage{url}
\makeatletter
\def\Uslash{\mathbin{\mathchar`\/}\@ifnextchar{/}{\kern-.15em}{}}
\g@addto@macro\UrlSpecials{\do \/ {\Uslash}}
\def\Ucolon{\mathbin{\mathchar`:}\@ifnextchar{/}{\kern-.1em}{}}
\g@addto@macro\UrlSpecials{\do : {\Ucolon}}
\makeatother

\begin{document}

\twocolumn[{\csname @twocolumnfalse\endcsname

\maketitle  

\begin{abstract}
\noindent
This paper presents a generalized framework for the simulation of multiple robots and drones in highly realistic models of natural environments. The proposed simulation architecture uses the Unreal Engine4 for generating both optical and depth sensor outputs from any position and orientation within the environment and provides several key domain specific simulation capabilities.  Various components and functionalities of the system have been discussed in detail.  The simulation engine also allows users to test and validate a wide range of computer vision algorithms involving different drone configurations under many types of environmental effects such as wind gusts.  The paper demonstrates the effectiveness of the system by giving experimental results for a test scenario where one drone tracks the simulated motion of another  in a complex natural environment.

\end{abstract}

\subsection*{Keywords}
Robot simulation, Drone simulation, Natural environment models, Natural feature tracking, \gls{UE4}.

\vspace*{1.0\baselineskip}
}]


\section{Introduction}

\copyrightspace
Graphical models of realistic natural environments are extensively used in games, notably simulation games and those that use immersive environments.  These virtual environments provide a high degree of interactive experience and realism in simulations.  Modern game engines provide tools for prototyping  realistic,  complex  and detailed virtual environments. Recently, this capability of game engines has been harnessed to the advantage of computer vision community to develop frameworks that can be used in scientific applications where vision based algorithms for detection, tracking and navigation could be effectively tested and evaluated with various types of sensor inputs and environmental conditions. This paper focuses on the development of a comprehensive standalone framework for multi-robot simulation  (specifically, multi-drone simulation) in complex natural environments, and proposes suitable configurations of tools, simulation architectures and also looks at key performance issues.

Several robot simulation engines exist which simulate different robots and vehicles e.g. multicopters, rovers, fixed wing  UAV, etc. Each engine has its advantages. The engines use large simulation environments consisting of models, sceneries, etc. generated by other simulation packages and frameworks. Following are some examples of such engines with dependencies on other simulation packages: 

\begin{itemize}
\item Standalone robot simulation engines using a flight simulator for models, sceneries and functions for visualization and simulation.  Examples of such packages are: (i) ArduPilot ~\cite{ardupilotfork} which communicates with Xplane ~\cite{xplane} and Flightgear ~\cite{flightgear} (ii) PX4 ~\cite{px4_firmware_fork} communicates with jmavsim ~\cite{jmavsim}. Flight simulators are usually much larger projects than robot simulation projects. They are more focused on user experience and interaction, but they also have visualization and dynamic simulation capabilities which are useful characteristics for drone projects. 

\item Standalone simulation packages that use physics engines, graphical interfaces and simulation capabilities provided by other simulation tools: for example PX4 ~\cite{px4_firmware_fork} with Gazebo ~\cite{gazebo}. Robot simulation environments are dedicated simulation environments. They are focused on giving proper tools for modeling and simulating robots but are less focused on visualization.  

\item Stand alone robot simulation environment: those environments include the robots and flying vehicle models. An example of that kind of environment is: \gls{morse}. Those environments are suitable for testing and evaluating ideas, but they don't have roots in real robot projects specifically in drone projects.

\item Game engine stand alone environment: the robot is simulated inside a game engine. For example a benchmark for tracking based on \gls{UE4} ~\cite{mueller2016benchmark}. Similarly to robot simulation environment, the drones inside game engines don't have roots in real drone projects. Additionally drone simulated in game engines don't share the dynamics of real drones. For example, they don't have to deal with wind gusts and vibrations.
\end{itemize}

In this paper, we propose a novel configuration that use game engines for the simulation environment, the primary motivation being the enhanced capabilities of a game engine such as UE4 in providing highly realistic environments and various modes of visualization. One of the primary advantages of this type of a configuration is that a game engine such as \gls{UE4} can provide realtime videos of camera output based on the position and attitude information of the robot.
This paper also gives an overview of the \gls{DRONESIMLAB}\ifanonymous\footnote{The name of the online project was replaced for anonymity}\fi ~\cite{dronelab} developed by us, which has constantly evolved with the analysis of various requirements and concepts related to the simulation architecture presented later in this work.  The design and implementation aspects  of the key components of this simulation engine have been presented in detail.

This paper is organized as follows: in section \ref{dronelab_section} we give an overview of the \gls{DRONESIMLAB} project. In section \ref{related work} we describe previous related work with game engines. Section \ref{section sim arch} gives detailed information about the framework simulation architecture and design goals. Section \ref{engine modifications} focuses on modifications needed to be made to meet the simulation design goals. Section \ref{performance} describes experimental results and performance. Section \ref{conclusions} provide information on future research directions as well as references to online demos of this research.

\section{The \gls{DRONESIMLABL} Project} \label{dronelab_section}
We developed \gls{DRONESIMLAB} as an opensource project to foster collaborative development of drone simulation packages that use the power and capabilities of the \gls{UE4} as discussed in the previous sections.  Some of the main functionalities which the current implementation provides are:
\begin{itemize}
\item  Multi-robot - can handle more than one robot and create visual interaction.
\item  \gls{sitl} driven - can simulate two drone models: ArduPilot and PX4 
\item  Based on Game Engine - Uses \gls{UE4} as an optical and depth sensor  
\item  Realtime - depends on the hardware but can run at 30 fps.
\item  Natural environments - can simulate trees, wind grass, etc. (comes with Game Engines assets).
\end{itemize}


\section{Related Works} \label{related work}

Some image processing and image-based algorithms have already been integrated into game engine/image simulation environments, although using game engines dedicated for games (like \gls{UE4}, CryEngine, and Unity) in simulations is much less common. We believe the reason for that is due that they are more focused on game experience as opposed to any real-world scientific applications involving simulations with associated mathematical and physical models and computer vision algorithms.

One example of such an approach is the autonomous landing of a \gls{vtol} \gls{uav} on a moving platform using image based visual servoing \cite{serra2014landing}, where they used gazebo simulation simultaneous localization and mapping. \gls{slam} \cite{afanasyev2015ros} is also tested and developed for indoor scenarios using gazebo simulation \cite{gazebo} \cite{meyer2012comprehensive}. Some environments are combined to create a more powerful engine. For example, \gls{morse} \cite{echeverria2011modular} combined with BGE (blender game engine) \cite{degroote:hal-01232114} and JSBsim (an open source flight dynamics model) 

  Lately, game engines have been increasingly used for simulations. Successful attempts have been made to evaluate the stability of structure using \gls{UE4}, creating photo-realistic scenes of stacks of blocks and applying deep learning methods \cite{DBLP:journals/corr/LererGF16}. A series of towers made from wooden cubes were created in a simulated environment using \gls{UE4} \cite{ue4}. Some of the towers were stable structures, and some collapsed when the dynamic simulation was run. A network was trained to detect the outcome of the experiments.  Testing the network on real environments achieved equal performance compared to human subjects in predicting whether the tower will fall. The most important aspect of this research is the fact that they could train the network on 180,000 scenarios which seems not feasible in a real life environment. 

 A more recent work connected \gls{UE4} with OpenCV \cite{opencv_library}, the project is called UnrealCV \cite{qiu2016unrealcv}. It extends the \gls{UE4} with a set of commands to interact with the virtual world. Another work \cite{mueller2016benchmark} proposed a new aerial video dataset and benchmark for low altitude \gls{uav} target tracking, as well as a photorealistic \gls{uav} simulator that can be coupled with tracking methods. Skinner \cite{skinner2016high} proposed a high-fidelity simulation for evaluating robotic vision performance for repeating robotic vision experiments under identical conditions. Similarly, we are providing a sandbox for high-fidelity simulations not only for algorithms but also for full \gls{sitl} simulations.  
 
Recently Microsoft released AirSim \cite{MSR-TR-2017-9}, an open source simulator based on Unreal Engine for autonomous vehicles from Microsoft AI \& Research which has a similar architecture as our proposed architecture. In their released implementation they are using their physics engine and control libraries.

\begin{figure}[H]
\centering
\centering
\resizebox{0.42\textwidth}{!}{\input{system_arch.tex}}
\caption{Simulation architecture}
\label{fig:figure_sim_arch}
\end{figure}
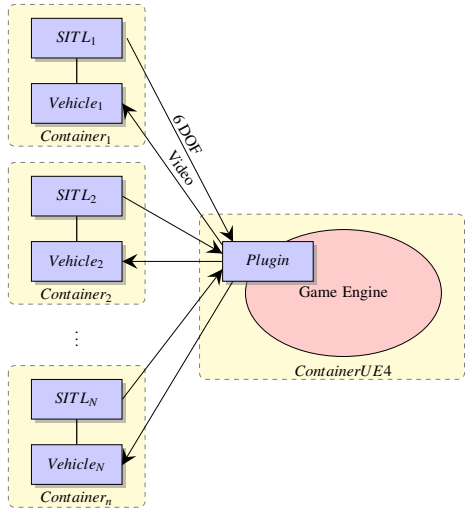

\section{SIMULATION ARCHITECTURE} \label{section sim arch}
In this paper, we propose a simulation architecture designed to meet the following primary goals:  (i) ability to generate realtime camera outputs for any arbitrary position and orientation in a natural environment, (ii) ability to integrate software and hardware in the loop simulations (iii) ability to combine multiple simulations and (iv) ability to reproduce results.  These aspects are elaborated below.

\subsection{Domain Specific Simulation Engine}
We focused on three simulation engines for the framework.
\begin{itemize}
\item The game engine provides video, depth data and additional visual environmental effects like wind and dust.
\item The physical model engine, usually supplied by the robot development framework. 
\item Supplimental simulation objects like communication channels models , computation power restrictions and additional simulation filters (for example a lens distortion filter). 
\end{itemize}

Engines can create an environment for a single robot. For instance in the case of \gls{sitl},  the simulation engine interacts with only one vehicle and produces sensory information for only one robot. On the other hand, the game engine can provide visual information for multiple robots as described in Figure ~\ref{fig:figure_sim_arch}, this is especially necessary if a visual interaction exists, for example; one robot can block the field of view for another robot, and this aspect should be implemented in the simulation. By embedding \gls{sitl} into our framework we ensure that the simulation is highly correlated with the real robot architecture  (since it is used for the robot development). \gls{hil} can be later used for further validation. 

\subsection{Simulated sensor architecture}
We identified three types of simulated sensors that can be used.
\begin{itemize}

\item  Single domain sensor - lives only in one engine. For example a simulated RGB camera from  UR4 ~\cite{ue4}. Another example is the  gyro sensor, which  is simulated only in the \gls{sitl}  software  e.g. gazebo, jmavsim, jsbsim etc.
\item  Multi domain  sensor - lives in more then one engine. For example such a sensor can be seen  in Figure ~\ref{fig:figurelabel1}.  In this example the simulated distance sensor gets information in from various sources like an external \gls{dem}.
\item  Complex sensor - lives in both the physical domain and in the simulated domain. An example of such sensor is a camera in front of a screen. The display provides the visual information and the camera is used just as in the real system, enabling monitoring real system performance and hardware issues. This concept is an extension of the HIL mode which combines hardware testing and software testing.
\end{itemize}
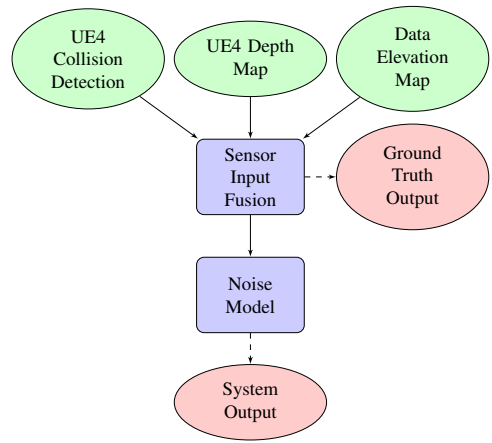
\begin{figure}[H]
\centering
\resizebox{0.42\textwidth}{!}{\input{sensor_arch.tex}}
\caption{Example of a multi-domain distance sensor, The simulated sensor accepts inputs from different engine resources and produces ground truth information and noisy output.}
\label{fig:figurelabel1}
\end{figure}

\subsection{Containers as vehicles}
We used a container approach to combine simulations not originally intended to work alongside each other. The container approach enables us to use different operating systems and libraries on the same machine. We can also control the network configuration in each container. This approach is different from other frameworks like 
AirSim \cite{MSR-TR-2017-9} by maintaining the original firmware developed by the Robot Simulation Framework. Our architecture can utilize the benefits of new features and continuing development of those environments. 

\subsection{Reproducibility}
The ability to reproduce results under differing or constant conditions is vital in system development as well as in research, and becomes more and more difficult as the complexity of the system increases. To realize this concept, we are using several existing software tools. We are using the Docker engine to manage system configuration, and Git version control to handle the software development. Since the experiments are in a simulation, other reproducibility aspects such as high-fidelity are built-in. In section \ref{performance}, we demonstrate testing of algorithms in complex natural environments by controlling simulation parameters. In this simulation, we can see that outdoor natural environment can be problematic for testing visual algorithms since we don't have full control of the environment. It seems that true reproducibility in an outdoor natural environment may be achieved only in simulation \cite{skinner2016high}.    

\subsection{Build system \& configuration management}
The simulation environment uses these software tools:
\begin{itemize}
\item Version Control - All files of this project are managed by Git version control under GitHub servers ~\cite{git}. The only exceptions are the \gls{UE4} projects which are managed locally due to the large file sizes. The UE4 ~\cite{ue4} source code is still managed by git in dedicated GitHub repository. For the purpose of sending realtime ground truth position, changes have been made both to ArduPilot Project and to PX4 and are managed in separate forks. Those changes are not compatible with the design and purpose of the original projects. Changes that were compatible (e.g. a turbulence model) were returned to the community as pull requests and then pulled back into our local fork.
\item Containers -  Created with Docker engine. 
\item ArduPilot Fork ~\cite{ardupilotfork} (Drone Project)
\item ROS - Supporting firmware for the PX4 project.
\item PX4 Fork ~\cite{px4_firmware_fork} (Drone Project)
\item \gls{UE4PYSERVER}\ifanonymous\footnote{The plugin name was replaced for anonymity}\fi ~plugin ~\cite{ue4pyserver}
\end{itemize}

\section{Engine Modifications} \label{engine modifications}
\subsection{\gls{UE4} Plugin}
Game engines are not dedicated research tools, obviously, but conveniently for our usage scenario they supply mechanisms like plugins to extend the capabilities of the engine. The plugin we used for the \gls{UE4} \cite{ue4} is called \gls{UE4PYSERVER} ~\cite{ue4pyserver} Plugin and was developed for the purpose of this research.
The main concepts behind the plugin development were:
\begin{itemize}
\item  Realtime: For this simulation, we took advantage of the realtime capabilities of the game engine. Realtime simulation (RT) is important when you want to run many tests in a short period.  RT simulation is also necessary when human interaction is involved because users expect realtime or near realtime behavior. To maintain RT behavior, the \gls{UE4} plugin was developed with minimal processioning on the  \gls{UE4} side. The primary purpose of the plugin is to communicate with other parts of the simulation. e.g. receiving 6 DOF information and sending video data.

\item Multi-Robot support - UE4 enables capture of the viewable screen to a file or a buffer, but this provides us with only one camera feed. To allow multiple cameras in the simulation, we used rendering-to-texture technique with object ScreenCapture2D ~\cite{screencapture2d}. The method is used in the game engine usually to render surfaces like security cameras, billboards, mirrors, etc. We used it to simulate a camera robot and depth sensors using the depth map provided by the ScreenCapture2D Object.
\item Synchronization -  We wanted the sampling to be synchronized for all the visual objects in the simulation. It is an important concept and might be critical for some applications, for example, simulating stereo camera.  For this purpose, we used coroutines which are a light version of synchronized pseudo-threads.
\end{itemize}

\subsection{Building realistic environment inside game engine for computer vision}
\begin{figure}
\centering
\includegraphics[width=0.5\textwidth]{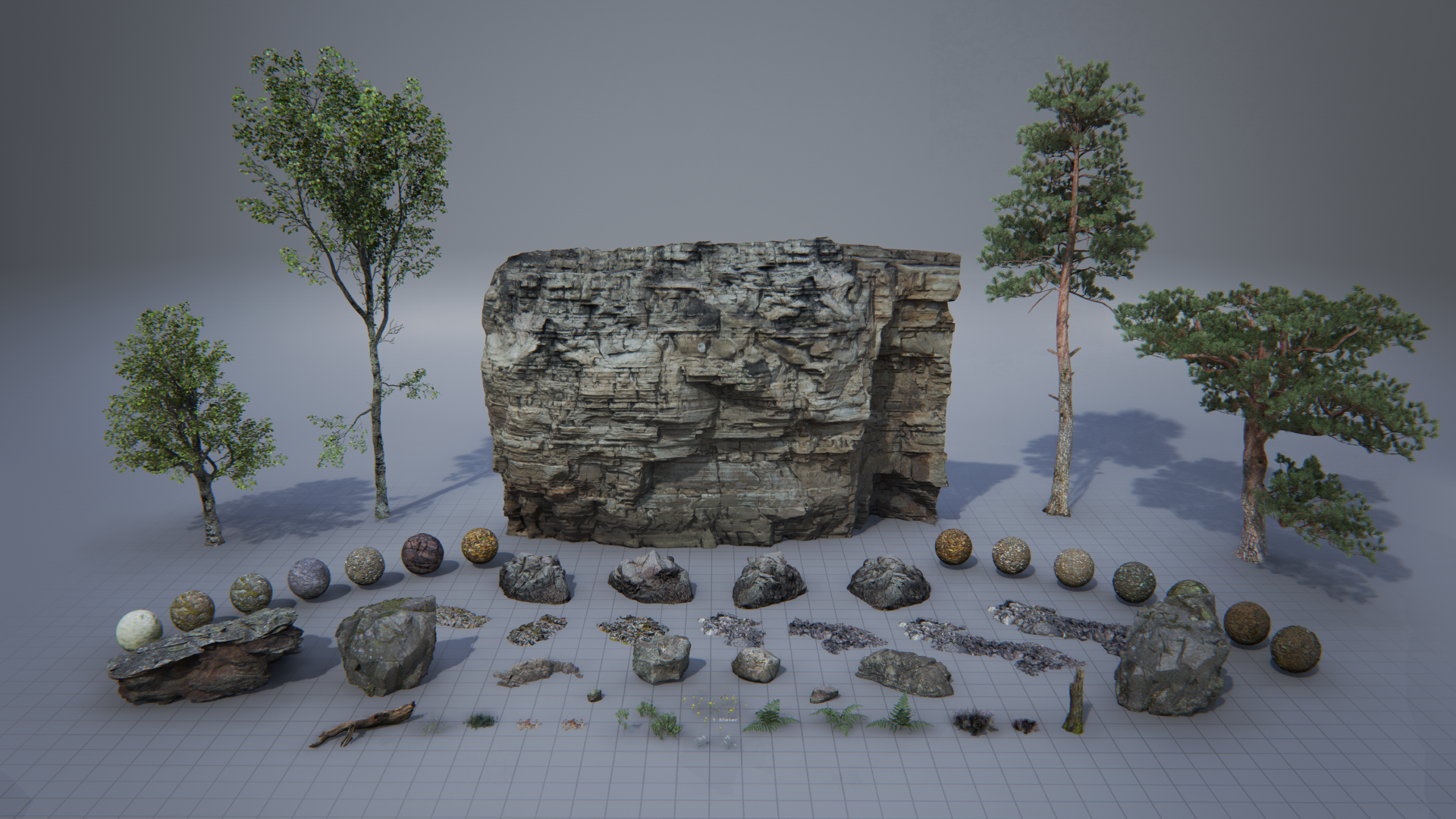}
\caption{Open world Overview - these are the assets used to build a forest environment}
\label{fig:openworldoverview}
\end{figure}

There are some special considerations when building virtual environments in game engines for computer vision purposes.
\begin{itemize}
\item \gls{lod} - in game engines using multiple \gls{lod} \cite[Chapter~3]{mooney2012unreal} in order to maintain graphics performance especially frame rate. This may create unnatural  textures changes which can be destructive for computer vision algorithms. For the purpose of this research, we can control the environment and the simulation, and we can use that to create a scene with only one \gls{lod}. 
\item Repeating patterns - In Figure \ref{fig:openworldoverview} we can see the meshes used to build the realistic scene. To reduce the effect of repeating patterns, each element is positioned in a different orientation and slightly different scaling. Also, the elements are positioned with some overlap with other objects which reduces the repeating effect.
\item Culling adjustments - the area rendered in the scene also known as frustum should be large enough for all the objects in the scene to be rendered, so we will avoid popping effects due to movements of the cameras or the objects themselves.
\item Dynamic shadows adjustments - moving objects in the scene like trees and robots should always cast dynamic shadows to imitate real scenarios.
\end{itemize}

\subsection{SITL}
The \gls{sitl} engine needs to send 6 DOF information at a high rate to the game engine (at least 30 fps) to maintain realtime constraints. For that purpose, some modifications are needed to the engine, so the \gls{sitl} engine will send ground truth information directly to the game engine, and also to logging mechanisms for later analysis as described in Figure \ref{fig:figure_sim_arch}.

\section{Experimental Results and Performance Analysis} \label{performance}
\subsection{Plugin tests in natural environment}
\begin{figure}
\centering
\includegraphics[height=2.5in]{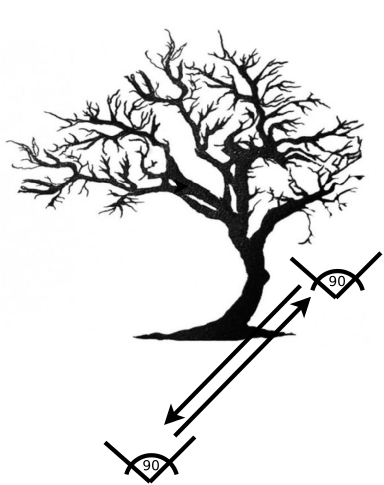}
\caption{Scene architecture - In the \gls{UE4} Editor, we placed a camera at an initial position, in front of a tree. We moved the camera diagonally away from the tree, and then returned to the same point. Camera maneuvers which starts and ends in the same position are ideal for tracking tests. Ideally, the tracked points should get back to the same original coordinates. }
\label{fig:scene arc}
\end{figure}

\begin{figure}
\centering
\begin{minipage}{.5\textwidth}
\centering
\includegraphics[height=2in]{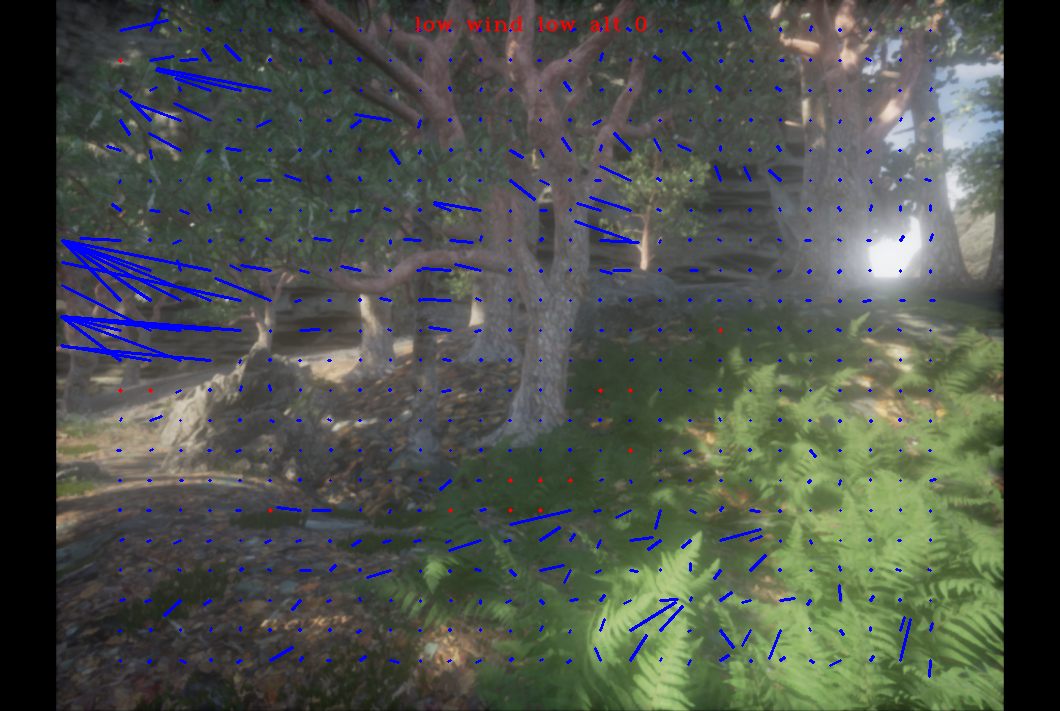}

(a)
\vspace{4ex}
\end{minipage}%

\begin{minipage}{.5\textwidth}
\centering
\includegraphics[height=2in]{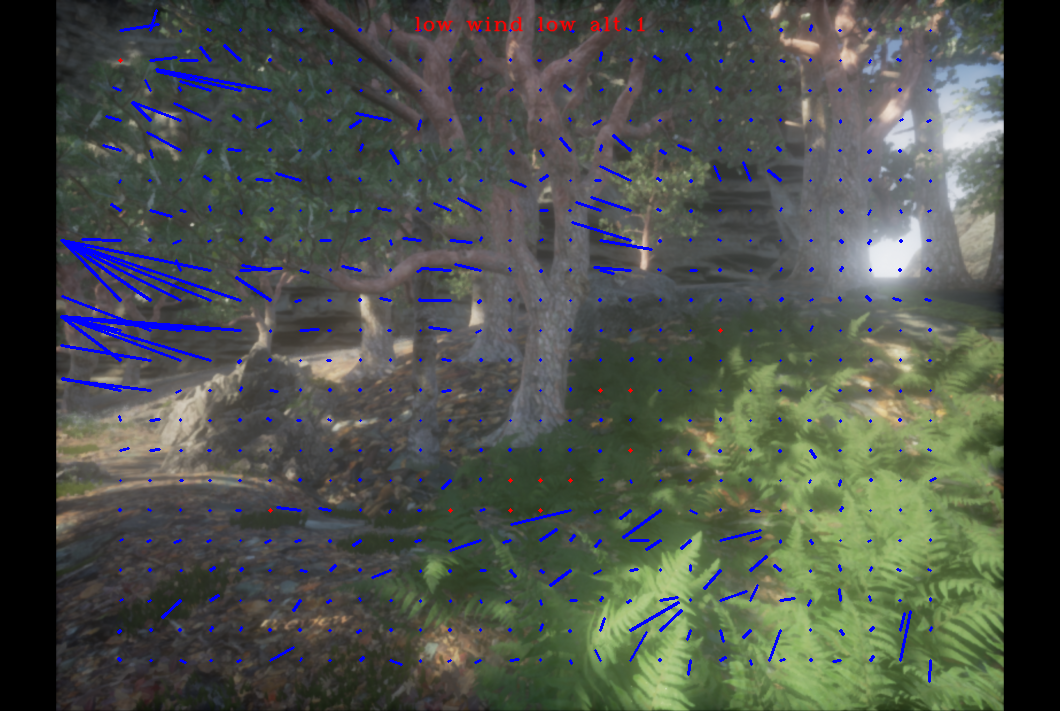}

(b)
\end{minipage}
\caption{Simulation outputs (a) and (b) showing reproducability of results under similar environmental conditions.}
\label{fig:similar conditions}
\end{figure}

\begin{figure}
\centering
\begin{minipage}{.5\textwidth}
\centering
\includegraphics[height=2in]{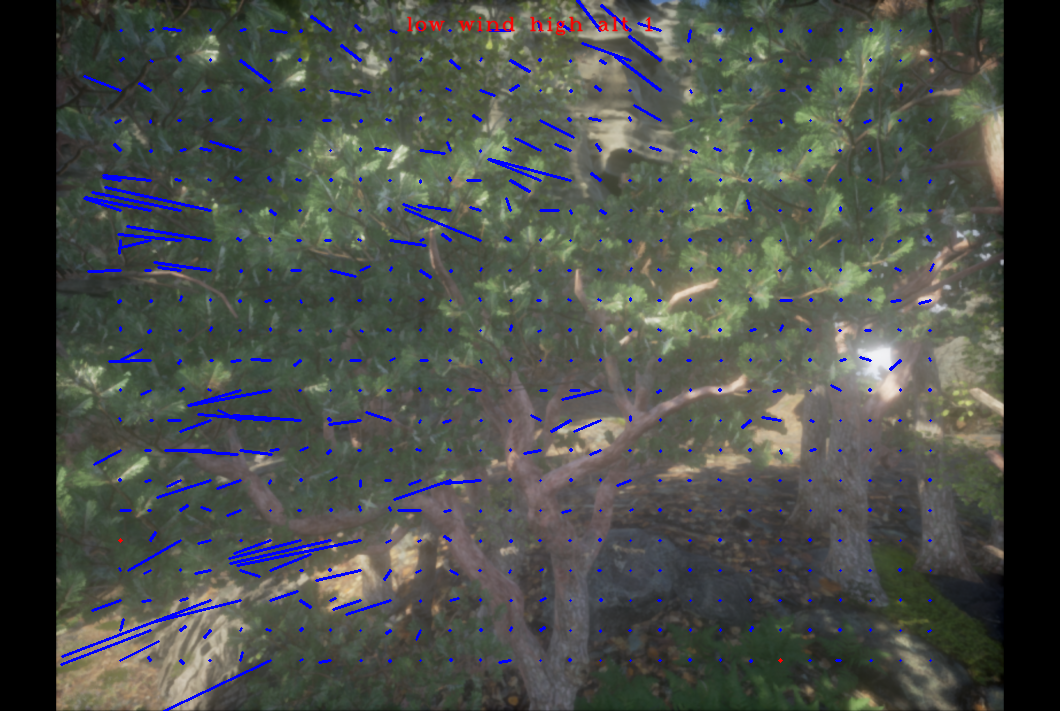}

(a)
\vspace{4ex}
\end{minipage}

\begin{minipage}{.5\textwidth}
\centering
\includegraphics[height=2in]{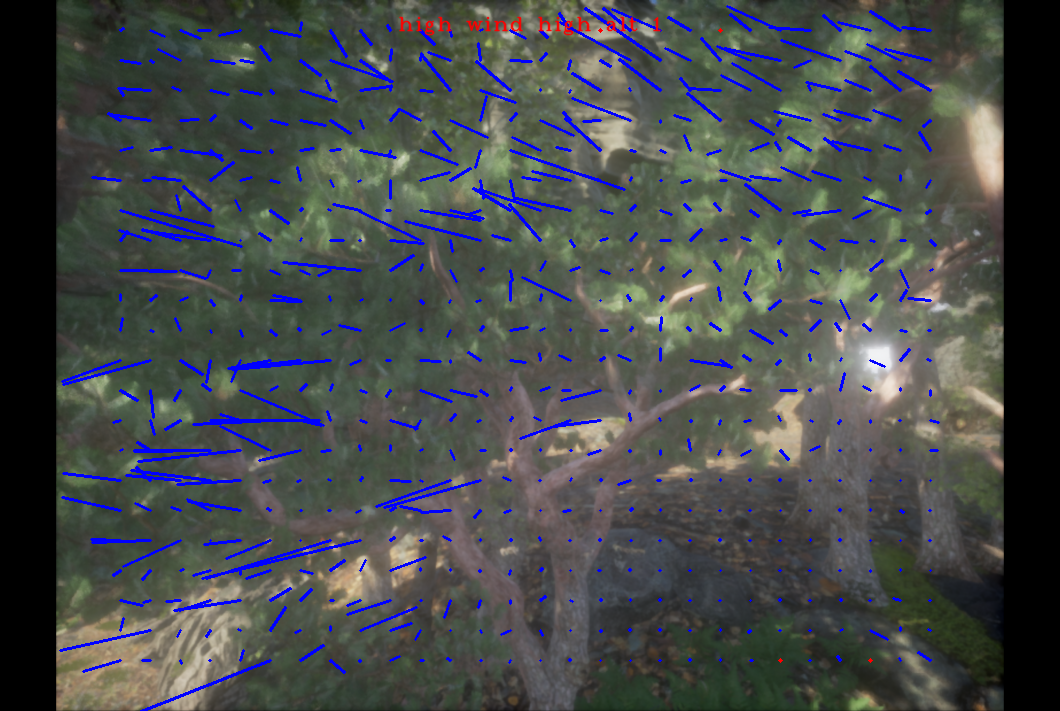}

(b)
\end{minipage}
\caption{Simulation outputs showing the variations in results under differing environmental conditions. The output in (a) was generated without simulated wind, while in (b), wind was added to the simulation.  We can observe degradation in the tracking quality of the grid features.}
\label{fig:different conditions}
\end{figure}

Running visual algorithms in a natural environment can be very challenging. Relative to artificial environments, natural scenes can by highly dynamic due to atmospheric conditions such as wind, and usually will not have distinct characteristics like straight lines, circles corners, etc. Using \gls{UE4PYSERVER} ~\cite{ue4pyserver} (which was developed as part of the simulation framework) and UE4 ~\cite{ue4} we developed a tracking simulation (live video can be found here ~\cite{pyserver_opencv}) to demonstrate the uniqueness of natural environment. The simulation is based on the Lucas-Kanade Optical Flow tracker implemented in the OpenCV library ~\cite{bouguet2001pyramidal} which we use it to track an ordered grid of points (no feature extraction). The maneuver is a simple camera facing forward and moving diagonally back and then return to the original position as described in Figure \ref{fig:scene arc}. Ideally, we would expect that the tracked points will return to the same coordinates when the simulation cycles back to the starting frame. Since this is a complex 3D scene, not all the points will return to the same location due to the loss of tracking, but in Figure \ref{fig:similar conditions} we can see that running the experiment twice produces similar results. Similar but not exact, since there is still some randomness in the scene due to movement of leaves that might cause slight differences. In Figure \ref{fig:different conditions} we conducted two experiments with the same setup, but in the second test, we add the wind to the scene by adding to the \gls{UE4} a Wind Direction Force object. We can see that the results are now very different. 
We repeated the experiment under various conditions and calculated the following MSE grade to quantify the tracking quality:
\begin{equation}
G = {\frac{1} {N}{\sum\limits_{P \in T_p } {|P_e - P_s|^{2} }}}  
\end{equation}
where: \(G\) is the tracking error, \(T_p\) is the tracked points, \(P_s\) an \(P_e\) are the start and end coordinates of the tracked points and \(N\) is the number of tracked points. Summarized results are in the following table: 

\begin{table}[htb]
\label{table:table1}
        \centering
        \begin{tabular}{|l|l|l|}
        \hline
         & low-wind & high-wind \\
        \hline
        \multirow{3}{*}{low-alt}  
        		& 96.36 (98.54\%) & 219.99 (97.40\%) \\
 				& 81.15 (98.54\%) & 225.48 (98.05\%) \\
 				& 87.69 (98.54\%) & 234.41 (97.89\%) \\
        \hline
        \multirow{3}{*}{high-alt}  
        		& 55.72 (97.56\%) & 243.59 (98.54\%) \\
 				& 53.82 (98.21\%) & 757.61 (97.73\%) \\
 				& 48.94 (97.40\%) & 382.78 (98.38\%) \\
        \hline
        \end{tabular}
        \caption{Tracking error (MSE) values in $pixels^2$. The numbers in brackets give the percentage of correctly tracked points.}
\end{table}
In Table 1, we can see the behavior of the tracking algorithm under different environmental conditions. As expected in high altitude (near the tree tops) with the combination of strong wind will be the most challenging scenario. As seen in the first column, the tracking error under low wind conditions is larger at low altitudes compared to high altitudes due to the presence of a higher density of objects such as leaves and branches that occlude the camera view at low altitudes.  On the other hand, when the wind speed increases, the trend is reversed because leaves and branches tend to move more than the objects closer to the ground like tree trunk rocks, etc.

\begin{figure}
\centering
\includegraphics[width=0.5\textwidth]{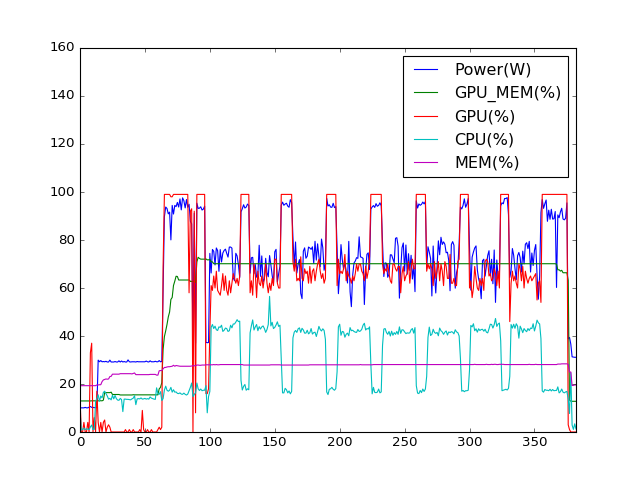}
\caption{Profiling scene - high level profiling during scene playback created with nvidia-smi and Python psutil.}
\label{fig:highlevelprofile}
\end{figure}

We developed a tool for profiling and monitoring the framework in addition to the existing tools in the \gls{UE4} editor. This is a high-level profiling tool that gives us the summary of the system utilization. An example of a test case is presented in Figures \ref{fig:scene arc} and \ref{fig:highlevelprofile}. fig ~\ref{fig:scene arc} explains the scene architecture and Figure \ref{fig:highlevelprofile} the corresponding profiling graph. The peaks in the GPU utilization are due to camera IO-intensive movements as would be expected. In this example, the GPU peaks in the graphs correspond to camera movement. When the camera is moving, we can see that the GPU is fully utilized (it reaches 100\%) resulting in reduced framerate and when the frame rate is reduced the CPU was less occupied because it was processing the images at a lower frame rate.

\subsection{\gls{DRONESIMLAB} tests}
\begin{figure}
\centering
\includegraphics[width=0.48\textwidth]{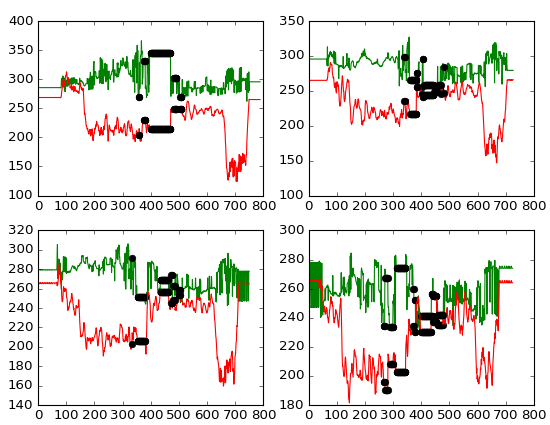}
\caption{Tracking results - Four consecutive tracking experiments results. The black dots represents tracking failures. The X axis is frame number and the Y axis is pixel position. Green and red lines are the X,Y pixel coordinates respectively.}
\label{fig:track_results}
\end{figure}

\begin{figure}
\centering
\includegraphics[width=0.4\textwidth]{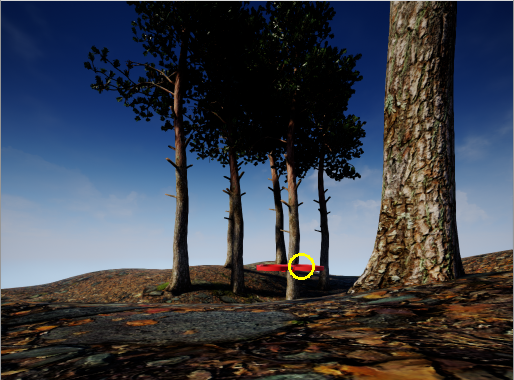}
\caption{Tracking Experiment - A drone following another drone in \gls{DRONESIMLAB}.}
\label{fig:track_image}
\end{figure}

We created a setup in \gls{DRONESIMLAB} for an experiment of one drone tracking another drone. The drones are Ardupilot drones simulated using their internal \gls{sitl} engine. We simulated the wind in the \gls{UE4} as well as in the \gls{sitl} engine including wind gusts. The two drones fly into the forest and then return to the original position \cite{hsvdronetrack1}. One of the drones is using HSV tracker \cite{hsvtrack} to track the other drone  (Figure \ref{fig:track_image}). We repeated this experiment four times and the results are presented in Figure \ref{fig:track_results}.

In all four scenarios, we can observe the loss of tracking capabilities when the drones enter the forest (black dots between frames 300 to 500 in all four scenarios). When the drones enter the woods, the shades from the forest canopy affects the color and brightness components of the drone as can be seen in this demo video \cite{hsvdronetrack1}. The threshold for tracking is not updated dynamically to demonstrate this behavior.
Other interesting phenomena is the high frequencies observed in the graph. These high frequencies result from the continuous maneuvering and changes in the 3D orientation of the drone to compensate for the high wind forces, which in turn results in variations in the estimation of the drones center position. In the last two experiments, we can see especially large amplitude in the beginning and at the end of the experiments as a result of the takeoff and landing process which provided different angles of viewing of the drone body led to a different estimation of the drone center.   

The above results have clearly demonstrated the usefulness of a game engine in not only producing realistic natural environments and their camera outputs, but also providing the ability to add and modify realistic environmental effects such as changes in wind parameters and illumination conditions.  These features allow us to generate ground truth data for various test conditions and to evaluate machine vision algorithms.

\section{CONCLUSIONS} \label{conclusions}
The creation of visually realistic environments is a very powerful tool for computer vision research as can be seen in section \ref{performance} and this corresponding video demo ~\cite{pyserver_opencv}. The \gls{DRONESIMLAB} project ~\cite{dronelab} aims to be a tool which adds game engines capabilities to the current existing robot simulation environments. The current work mainly focuses on \gls{UE4}, but adding another game engine may increase the dimensionality of modifiable parameters in our systems. For instance, training deep learning algorithms on multiple worlds each created by a different game engine may more accurately generalize to the real world domain. This paper has presented a new framework for simulating multi-robot (specifically, multi-drone) motion in such environments, where environmental effects can be easily incorporated, and complex computer vision tasks evaluated. The simulation architecture along with the key functionalities of the simulation engine have been discussed in detail.

\ifanonymous
\bibliography{paper,anonymous}{}
\else
\bibliography{paper,myref}{}
\fi
\bibliographystyle{plain}

\end{document}

%% file: paper_gls.tex
\newglossarystyle{clong}{%
    {\tablehead{}\tabletail{}%
     \begin{supertabular}{lp{\glsdescwidth}}}%
    {\end{supertabular}}%
  \renewcommand*{\glsgroupheading}[1]{}%
}

\newacronym{sitl}{SITL}{Software In The Loop}

\newacronym{UE4}{UE4}{Unreal Engine 4}

\newacronym{hil}{HIL}{Hardware In The Loop}

\newacronym{morse}{MORSE}{Modular Open Robots Simulation Engine}

\newacronym{slam}{SLAM}{Simultaneous Localization and Mapping}

\newacronym{lod}{LOD}{Level of Details}

\newacronym{uav}{UAV}{Unmanned Aerial Vehicle}

\newacronym{vtol}{VTOL}{Vertical Takeoff and Landing}
\newacronym{dem}{DEM}{Digital Elevation Model}

\ifanonymous
\newglossaryentry{DRONESIMLABL}{%
name={PROJECTX},%
description={}%
}

\newglossaryentry{DRONESIMLAB}{%
name={ProjectX},%
description={}%
}
\newglossaryentry{UE4PYSERVER}{
name={ServerY},%
description={}%
}
\else
\newglossaryentry{DRONESIMLAB}{%
name={DroneSimLab},%
description={}%
}
\newglossaryentry{DRONESIMLABL}{%
name={DRONESIMLAB},%
description={}%
}
\newglossaryentry{UE4PYSERVER}{
name={UE4PyServer},%
description={}%
}

\fi

%% file: system_arch.tex
\pgfdeclarelayer{background}
\pgfdeclarelayer{foreground}
\pgfsetlayers{background,main,foreground}

\tikzstyle{sensor}=[draw, fill=blue!20, text width=5em, 
    text centered, minimum height=2.5em,drop shadow]
\tikzstyle{ann} = [above, text width=5em, text centered]
\tikzstyle{wa} = [draw, ellipse,fill=red!20, text width=8em, text centered, minimum height=8em]

\tikzset{myptr1/.style={decoration={markings,mark=at position 1 with %
    {\arrow[scale=3,>=stealth]{>}}},postaction={decorate}}}
\tikzset{myptr2/.style={-,thick}}

\def\blockdist{2.3}
\def\edgedist{2.5}
\begin{tikzpicture}
    \node (aaa) [wa]  {Game Engine};
	\path (aaa.west)+(0.5,0.7) node (wa)[sensor] {$Plugin$};    

    \path (wa.west)+(-3.2,3.5) node (asr1) [sensor] {$Vehicle_1$};
    \path (asr1.north)+(0,+1) node (sitl1) [sensor] {$SITL_1$};
    \path (wa.west)+(-3.2,-0.0) node (asr2)[sensor] {$Vehicle_2$};
    \path (asr2.north)+(0,+1) node (sitl2) [sensor] {$SITL_2$};

    \path (wa.west)+(-3.2,-2.0) node (dots)[ann] {$\vdots$}; 

	\path (wa.west)+(-3.2,-4.5) node (asrn)[sensor] {$Vehicle_N$};    
    \path (asrn.north)+(0,+1) node (sitln) [sensor] {$SITL_N$};

    \draw [myptr1] (sitl1.east) + (0.1,0) -- (wa.150) {} (wa.150); 
    \path (sitl1.east) -- (wa.150) node [midway, above, sloped] () {6 DOF};
    \draw [myptr1] (wa.160) -- (asr1.east) {} (asr1.east); 
    \path (asr1.east) -- (wa.160) node [midway, above, sloped] () {Video};
	\draw [myptr2] (asr1.north) -- (sitl1.south) ;

    \draw [myptr1] (sitl2.east) -- (wa.170) {} (wa.170); 
    \draw [myptr1] (wa.180) -- (asr2.east) {} (asr2.east); 
	\draw [myptr2] (asr2.north) -- (sitl2.south) ;

    \draw [myptr1] (sitln.east) -- (wa.190) {} (wa.190); 
    \draw [myptr1] (wa.210) -- (asrn.east) {} (asrn.east); 
	\draw [myptr2] (asrn.north) -- (sitln.south) ;

    \path (asr1.south) +(0,-0.3) node (asrs1) {$Container_1$};
  
    \begin{pgfonlayer}{background}
        \path (sitl1.west |- sitl1.north)+(-0.5,0.3) node (a) {};
		\path (asr1.east |- asr1.south)+(+0.5,-0.5) node (c) {}; 
        \path[fill=yellow!20,rounded corners, draw=black!50, dashed]
            (a) rectangle (c);                       
    \end{pgfonlayer}

    \path (asr2.south) +(0,-0.3) node (asrs2) {$Container_2$};

    \begin{pgfonlayer}{background}
        \path (sitl2.west |- sitl2.north)+(-0.5,0.3) node (a) {};
		\path (asr2.east |- asr2.south)+(+0.5,-0.5) node (c) {}; 
        \path[fill=yellow!20,rounded corners, draw=black!50, dashed]
            (a) rectangle (c);                       
    \end{pgfonlayer}

    \path (asrn.south) +(0,-0.3) node (asrsn) {$Container_n$};

    \begin{pgfonlayer}{background}
        \path (sitln.west |- sitln.north)+(-0.5,0.3) node (a) {};
		\path (asrn.east |- asrn.south)+(+0.5,-0.5) node (c) {}; 
        \path[fill=yellow!20,rounded corners, draw=black!50, dashed]
            (a) rectangle (c);                       
    \end{pgfonlayer}

    \begin{pgfonlayer}{background}
        \path (wa.west |- wa.north)+(-0.5,0.6) node (a) {};
		\path (aaa.east |- aaa.south)+(+0.5,-0.5) node (c) {}; 
        \path[fill=yellow!20,rounded corners, draw=black!50, dashed]
            (a) rectangle (c);                       
    \end{pgfonlayer}
	\path (aaa.south) +(0,-0.3) node (aaan) {$Container UE4$};

\end{tikzpicture}

%% file: sensor_arch.tex
\tikzstyle{decision} = [diamond, draw, fill=blue!20, 
    text width=4.5em, text badly centered, node distance=3cm, inner sep=0pt]
\tikzstyle{block} = [rectangle, draw, fill=blue!20, 
    text width=5em, text centered, rounded corners, minimum height=4em ,distance=4cm]
\tikzstyle{line} = [draw, -latex']
\tikzstyle{cloud} = [draw, ellipse,fill=green!20, node distance=3cm, text width=5em, text centered, minimum height=4em]
\tikzstyle{cloudr} = [draw, ellipse,fill=red!20, node distance=3cm, text width=5em, text centered, minimum height=4em]
\tikzstyle{cloudr1} = [draw, ellipse,fill=red!20, node distance=2cm, text width=5em, text centered, minimum height=4em]
    
\begin{tikzpicture}[node distance = 2.2cm, auto]
    \node [cloud] (ue4depth) {UE4 Depth Map};
    \node [cloud, left of=ue4depth] (ue4cold) {UE4 Collision Detection};
    \node [cloud, right of=ue4depth] (dem) {Data Elevation Map};
    \node [block, below of=ue4depth] (Fusion) {Sensor Input Fusion};
    \node [cloudr, right of=Fusion] (ground_truth) {Ground Truth Output};
    \node [block, below of=Fusion] (Noise) {Noise Model};
    \node [cloudr1, below of=Noise] (system) {System Output};
    \path [line] (ue4depth) -- (Fusion);
    \path [line] (ue4cold) -- (Fusion);
    \path [line] (dem) -- (Fusion);
    \path [line] (Fusion) -- (Noise);
    \path [line,dashed] (Fusion) -- (ground_truth);
    \path [line,dashed] (Noise) -- (system);
\end{tikzpicture}